\documentclass[letterpaper, 10 pt, conference]{ieeeconf}  

\IEEEoverridecommandlockouts                              

\usepackage{lipsum}
\usepackage{amsmath, amsfonts, amssymb}
\usepackage{stmaryrd}
\usepackage{mathrsfs}
\usepackage{bbold}
\usepackage{graphicx}
\usepackage{color, soul}
\usepackage{xcolor}
\usepackage{siunitx} 
\usepackage{tabularx}
\usepackage{booktabs}
\usepackage{marginnote}
\usepackage{tablefootnote}
\usepackage{graphicx,subfigure,float}
\usepackage[font=small,skip=0pt,belowskip=-10pt]{caption}
\usepackage{bm}
\usepackage{epstopdf}

\renewcommand{\baselinestretch}{0.985}

\newcommand*{\vect}[1]{\mathbf{#1}}
\newcommand*{\vectd}[1]{\dot{\mathbf{#1}}}
\newcommand*{\vectdd}[1]{\ddot{\mathbf{#1}}}

\newcommand*{\quat}[1]{\mathit{#1}}

\newcommand*{\dvq}[1]{\hat{\mathbf{#1}}}
\newcommand*{\dvqd}[1]{\dot{\hat{\mathbf{#1}}}}

\newcommand*{\dq}[1]{\hat{\mathit{#1}}}
\newcommand*{\dqd}[1]{\dot{\hat{\mathit{#1}}}}

\setstcolor{red}

\title{\LARGE \bf
Dual Quaternion Control of UAVs with Cable-suspended Load}

\author{Yuxia Yuan$^{1}$ and Markus Ryll$^{2}$
\thanks{$^{1}$Yuxia Yuan is with Autonomous Aerial Systems, School of Engineering and
Design, Technical University of Munich, Lise-Meitner-Straße 9, 85521 Ottobrunn, Germany
{\tt\small yuxia.yuan@tum.de,yuxia.yuan2@outlook.com}}%
\thanks{$^{2}$Markus Ryll is with Munich Institute of Robotics and Machine Intelligence (MIRMI) and Autonomous Aerial Systems, School of Engineering and Design, Technical University of Munich, Lise-Meitner-Straße 9, 85521 Ottobrunn, Germany
{\tt\small markus.ryll@tum.de}}%
}

\begin{document}

\maketitle
\pagenumbering{arabic}

\begin{abstract}
Modeling the kinematics and dynamics of robotics systems with suspended loads using dual quaternions has not been explored so far.
This paper introduces a new innovative control strategy using dual quaternions for UAVs with cable-suspended loads, focusing on the sling load lifting and tracking problems.  
By utilizing the mathematical efficiency and compactness of dual quaternions, a unified representation of the UAV and its suspended load's dynamics and kinematics is achieved, facilitating the realization of load lifting and trajectory tracking.
The simulation results have tested the proposed strategy's accuracy, efficiency, and robustness.
This study makes a substantial contribution
to present this novel control strategy that harnesses the benefits of dual quaternions for cargo UAVs. Our work also holds promise for inspiring future innovations in under-actuated systems control using dual quaternions.

\end{abstract}

\section{Introduction}
\label{sec:1 Introduction}
In recent years, interest in using aerial robots, commonly known as unmanned aerial vehicles (UAVs), across diverse applications has experienced fast growth.
The multifaceted capabilities of UAVs—vertical takeoff and landing, hovering with directional changes, and lateral and altitude adjustments—have opened an extensive array of applications, including surveillance, photography, rescue missions, cargo delivery, and so on~\cite{macrina2020review,omar2023reviewload}. 
Notably, the realm of cargo transportation has gained substantial popularity, evidenced by major online retailers like Amazon, Google, DHL, and Walmart announcing the incorporation of UAVs into their parcel delivery operations~\cite{macrina2020review}. 
Utilizing a cable to connect a UAV and its load is efficient and cost-effective, but it introduces greater control challenges for UAVs due to the potential swinging of the cargo, which can impact UAV dynamics~\cite{goodarzi2014geometric}. 
Consequently, a crucial aspect of UAV transportation control is to limit the swing and oscillations of the load suspended beneath the UAV.
In this paper, we use the term "cargo UAV" to denote a UAV with a cable-suspended load beneath it.

Controlling UAVs poses challenges due to the nonlinearity of their dynamics and the under-actuation. A UAV's control typically involves managing attitude and altitude through four thrust inputs. 
Researchers have made several efforts in developing control systems, such as proportional-derivative controllers, linear quadratic regulators (LQR), and nonlinear controllers~\cite{derrouaoui2022review}. 
Among these, a geometric controller proposed in~\cite{lee2010geometric} stands out. It expressed UAV dynamics globally on the configuration manifold of the special Euclidean group $SE(3)$. 
Compared to other geometric control approaches, it performed well in controlling an underactuated UAV. 
The reseachers then further extended their work on a cargo UAV system, defined on the configuration space $SE(3)\times 
 S2$~\cite{sreenath2013geometric}.
This geometric controller was then widely applied to cargo UAV control~\cite{goodarzi2014geometric,cruz2017geometric,Pries2021}.
For example, the authors proposed a hybrid model in~\cite{cruz2017geometric} to lift a cable-suspended load based on this geometric controller. They decomposed the lifting maneuver into three modes in this study: \textit{Setup}, \textit{Pull}, and \textit{Raise} to formulate the dynamics of the cargo UAV system separately. 

Further explorations into cargo UAV systems include LQR control algorithm~\cite{alothman2015lifting}, decoupling of cargo UAV dynamics in longitudinal and lateral planes~\cite{belguith2023lonlat}, and a fixed-time control approach to enhance system robustness considering external disturbances~\cite{lv2022fixed}. 
Additionally, researchers also trained neural networks for cargo UAV control, as demonstrated in~\cite{lopez2022cnn}, where a convolutional neural network was used to estimate the position and orientation of the load.
\begin{figure}[tp]
    \centering
    \includegraphics[width=7cm]{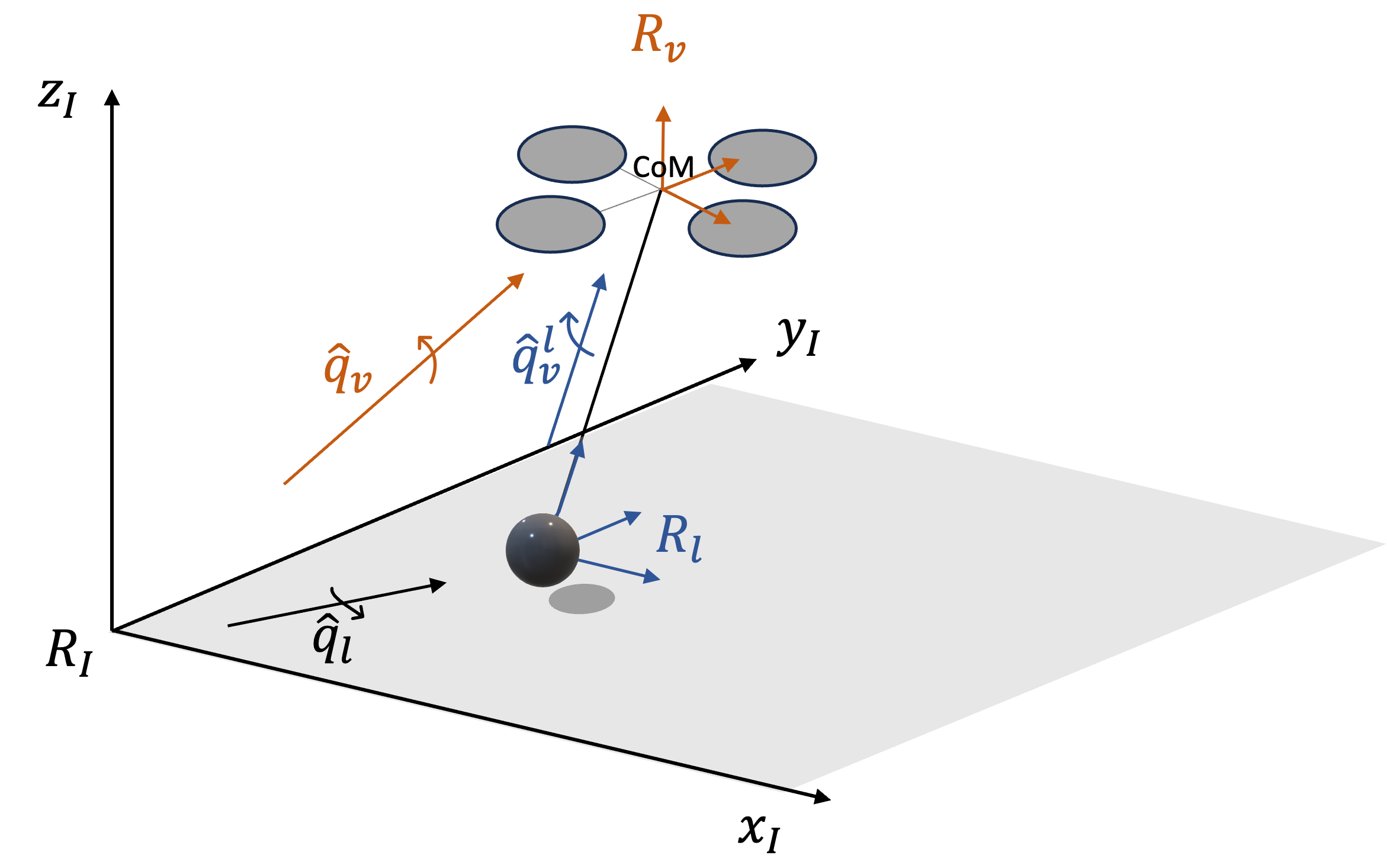}
    \caption{Dual quaternion transformations among three frames. $R_I$ is the inertial frame, $\dq{q}_v$ denotes the transformation from $R_I$ to $R_v$ frame, and $\dq{q}_l$ is the transformation from $R_I$ to $R_l$. $\dq{q}_v^l$ indicates the transformation from $R_l$ to $R_v$.}
    \label{fig:frames}
\end{figure}

This paper introduces a novel control strategy for cable-suspended UAVs using dual quaternions to enhance the efficiency and compactness of sling load lifting and trajectory tracking. Dual quaternion transformations between different frames are shown in Fig.~\ref{fig:frames}, which provides a mathematical framework that allows us to represent both the dynamics and kinematics of the UAV and its sling load in a unified representation. 
Dual quaternions can be seen as an extension of unit quaternions to represent the position and orientation of a rigid body~\cite{dantam2014quaternion}.
They provide several advantages compared to homogeneous transformation matrices~\cite{bullo1995HTM, lee2010geometric} for the representation of pose transformation, such as higher computational efficiency, mathematical compactness, lack of singularities, and optimal rigid transformation blending.
These features of dual quaternions have been used for the description of rigid-body kinematic models like the pose following control law in~\cite{arrizabalaga2023pose}, and the modeling of kinematics and dynamics of systems like UAVs~\cite{wang2013unit,abaunza2017dual}.
The authors of~\cite{wang2013unit} provided a trajectory tracking controller based on unit dual quaternions for a rigid body, where the error dynamics were described by a unit dual quaternion. 
\cite{abaunza2017dual} adopted a similar approach for quadrotor aerial manipulator control. The manipulator in their work also modeled an external load for the quadrotor, despite being connected using rigid links rather than a flexible cable. 

To the best of the authors’ knowledge, modeling the dynamics of UAVs with cable-suspended loads using dual quaternions has not been addressed in the literature so far.  
To explore and meet this research gap, this paper provides a novel control strategy to control cargo UAVs using dual quaternions. 
The cargo UAV system can be treated as two subsystems, the UAV and the sling-load, involving three orientation frames (the inertial frame, the sling-load frame, and the UAV frame). 

The main contributions of this study are twofold: First, we formulate the dynamic model for under-actuated cargo UAV systems using dual quaternions, which facilitates the analysis of the translational and rotational dynamics of a UAV with a sling load. 
Second, we develop a novel control strategy for UAVs with cable-suspended load to track a desired trajectory. The development of this controller is inspired by the unit dual quaternion tracker~\cite{wang2013unit} and geometric control~\cite{sreenath2013geometric}.
Finally, we tested our approach using two sets of simulations. The first one shows that our developed controller performs well for load lifting and trajectory tracking maneuvers. The second simulation evaluates the efficiency and robustness of our control strategy by considering system uncertainties and external disturbances and comparing it with the geometric controller. 
Our work not only addresses the technical challenges associated with cargo UAV control, but also opens up avenues for further innovation in the realm of cargo UAV control using dual quaternions.

The rest of this paper is structured as follows. Sec.~\ref{sec:2 Assumptions and Preliminary Knowledge} outlines the assumptions for the cargo UAV systems and introduces the mathematical basis of dual quaternions. 
The dynamic models of the cargo UAV system considering the cable state are presented in Sec.~\ref{sec:3 Cargo UAV System}.
Then, the control strategies to address lifting and tracking problems are stated in Sec.~\ref{sec:4 Control design}, along with the decomposition of the lifting maneuver into three modes.
In Sec.~\ref{sec:5 simulation}, the system setup and the simulation results are discussed. 
Finally, Sec.~\ref{sec:6 Conclusion} concludes our work.

\section{Assumptions and Preliminary Knowledge}
\label{sec:2 Assumptions and Preliminary Knowledge}
\subsection{Assumptions and Notations}
\label{subsec:assumptions}
The cargo UAV studied in our work consists of a UAV and a sling load. It can be considered as two subsystems and each of them is a rigid body, which involves three orientation frames, as shown in Fig.~\ref{fig:frames}. The load is physically attached to the UAV by a cable. Modeling all internal and external forces can quickly become very complex. For this study, we made the following simplifications and assumptions:
\begin{enumerate}
    \item We ignore elastic deformations of the rope, the UAV, and the load.
    \item The load is considered as a point mass.
    \item The cable is mass-free, and only forces and no torques are transmitted via the cable.
    \item The offset between the attached point of the load with the CoM of the UAV is considered as zero.
    \item The air drag and other disturbances are ignored.
\end{enumerate}

Before going further, the following notation rules are adopted:

$\begin{array}{ll}
_v, _l, _c & \text{Subscripts indicate UAVs, loads, and cables.} \\ 
^d         & \text{Superscript $d$ represents the desired states.} \\ 
^b         & \text{Superscripts $b$ indicate the body frame.} \\  
\quat{q}, \dq{q} \in \mathbb{H} & \text{Quaternions and dual quaternions.} \\
\vect{v} \in \mathbb{R}^{3}  & \text{Vectors (vector quaternions).} \\
\dvq{v} \in \mathbb{H} & \text{Dual vector quaternions.} \\
\quat{I}, \quat{0} \in \mathbb{H} & \text{Unit quaternion $[1,0,0,0]$ and zero quaternion.} \\
\vect{0} \in \mathbb{H} & \text{Zero vector quaternion.} 
\end{array}$

\subsection{Dual Quaternion}
As described in~\cite{dantam2014quaternion}, the components of a dual quaternion are both quaternions belonging to $\mathbb{H}$, which is represented as $\dq{q} = q_r + q_d \epsilon, q_r, q_d \in \mathbb{H}$.
The involved operations for dual quaternions are listed below:
\begin{itemize}
    \item Sum: $\dq{q}_1 + \dq{q}_2 = q_{1r} + q_{2r} + [q_{1d} + q_{2d}] \epsilon$
    \item Product: $\dq{q}_1 \otimes \dq{q}_2 = q_{1r} \otimes q_{2r} + [q_{1r} \otimes q_{2d} + q_{1d} \otimes q_{2r}] \epsilon$
    \item Conjugate: $\dq{q}^* = q_r^* + q_d^* \epsilon$
    \item Norm for unit dual quaternions: $||\dq{q}||^2 = 1 + 0\epsilon$
    \item Inverse for unit dual quaternions: $\dq{q}^{-1} = \dq{q}^*$.
    \item Natural logarithm mapping: $\ln \dq{q} = \frac{1}{2}(\overline{\theta} + T \epsilon)$; $\overline{\theta}$ is the rotation angle vector.
    \item Adjoint transformation: $Ad_{\dq{q}}\dq{r} = \dq{q} \otimes \dq{r} \otimes \dq{q}^*$
    \item Dual vector quaternion: $\dvq{v} = \vect{v}_r + \vect{v}_d \epsilon $. $\dvq{v}$ can be treated as $\dq{v} = [0, \vect{v}_r] + [0, \vect{v}_d] \epsilon $. 
\end{itemize}

\subsection{Dual Quaternion Dynamic Model}
\label{subsec:dual dynamics}
Using the Newton-Euler approach, the rotational dynamics of a rigid body can be expressed as
\begin{equation}
    J\vectd{\omega}^b + \vect{\omega}^b \times (J\vect{\omega}^b) = \vect{\tau}
    \label{eq:tau}
\end{equation}
where $J \in \mathbb{R}^{3 \times 3}$ is the inertia matrix of the body, and $\vect{\tau} \in \mathbb{R}^3$ is 
the sum of the control input torques and the external torques. Similarly, the translational dynamics relative to its body frame is given by
\begin{equation}
    m \vectdd{T}^b = \vect{F}
    \label{eq:F}
\end{equation}
where $m \in \mathbb{R}$ is the mass and $\vect{F} \in \mathbb{R}^{3}$ is
the sum of the control input forces and external forces (e.g., gravity) in the body frame. 

The kinematics of a rigid body in terms of its orientation, position, and twist is given in~\cite{wang2013unit} by
\begin{gather}
\label{eq:qdot}
    \dqd{q} = \frac{1}{2}\dq{q} \otimes \dq{\xi}^b \\
\label{eq:twist}
    \dq{\xi}^b = \vect{\omega}^b + [\vect{\omega}^b \times \vect{T}^b + \vectd{T}^b]\epsilon 
\end{gather}
where $\dq{\xi}^b$ is the twist, which is a dual quaternion combining rotational velocity $\vect{\omega}^b$ and translational velocity $\vectd{T}^b$.
Then the dual quaternion dynamic model of a rigid body is given by Eq~\eqref{eq:motion equations}. For details, readers can refer to~\cite{wang2013unit}.
\begin{equation}
\label{eq:motion equations}
    x = \left[ \begin{array}{c}
         \dq{q}  \\
         \dq{\xi}^b 
    \end{array} \right],~~ \dot{x} = \left[ \begin{array}{c}
         \dqd{q}  \\
         \dqd{\xi}^b 
    \end{array}\right] = \left[ \begin{array}{c}
         \frac{1}{2} \dq{q} \otimes \dq{\xi}^b \\
         \dq{F} + \dq{u} 
    \end{array}\right]
\end{equation}
with 
\begin{equation}
    \begin{cases}
        \dq{F} = a + [a \times \vect{T}^b + \vect{\omega}^b \times \vectd{T}^b] \epsilon \\ \nonumber
        \dq{u} = J^{-1} \vect{\tau} + [J^{-1} \vect{\tau} \times \vect{T}^b + m^{-1} \vect{F}] \epsilon \\ \nonumber
        a = -J^{-1}\vect{\omega}^b \times J\vect{\omega}^b \nonumber
    \end{cases}.
\end{equation}

\subsection{Dual Quaternion Error Dynamics}
\label{subsec:error dynamics}
Given the desired configuration $\dq{q}^d$, and the twist $\dq{\xi}^d$, the errors
can be found as
\begin{gather}
    \label{eq:q_e}
    \dq{q}_e = \dq{q}^{d*} \otimes \dq{q}, \\
    \label{eq:xi_e}
    \dq{\xi}^b_e = \dq{\xi}^b - Ad_{\dq{q}_e^{*}} \dq{\xi}^d
\end{gather}
where $\dq{q}^{d*}$ is the conjugate of $\dq{q}^d$ and $Ad_{\dq{q}_e^{*}} \dq{\xi}^{d}$ is adjoint transformation.
For calculation details, readers are referred to~\cite{wang2013unit}.  

By taking the derivative of $\dq{\xi}^b_e$, we get 
\begin{equation}
    \label{eq:xi_e_dot}
    \dqd{\xi}^b_e = \dqd{\xi}^b - E_{\dq{q}_e^{*}}\dq{\xi}^{d}
\end{equation}
with $E_{\dq{q}_e^{*}}\dq{\xi}^{d} = \dqd{q}_e^{*} \otimes \dq{\xi}^{d} \otimes \dq{q}_e + Ad_{\dq{q}_e^{*}} \dqd{\xi}^{d}  + \dq{q}_e^{*} \otimes \dq{\xi}^{d} \otimes \dqd{q}_e$. 

From Eq.~\eqref{eq:motion equations}, we have
\begin{equation}
    \label{eq:xi_e_dot2}
    \dqd{\xi}^b_e = \dq{F} + \dq{u} - E_{\dq{q}_e^{*}}\dq{\xi}^{d}.
\end{equation}

\section{The Cargo UAV System}
\label{sec:3 Cargo UAV System}
The cargo UAV system contains a sling load attached to the CoM of the UAV, which can be separated into two subsystems, the sling-load and the UAV, to study its dynamics. In the following, we develop cable state-dependent scenarios to derive the complete model.

\subsection{Dual Vector Quaternion of the Dynamics of the Sling-load}
\label{subsec:dq sling load}
In our research, the sling load is attached to the UAV by a cable. We treat the load and the cable as a sling-load subsystem. Though we assume the cable is rigid, there is no torque transmitted by the cable to the load. The dual vector quaternion $\dvq{q}$ is adopted to describe the configuration of the sling load as
\begin{gather}
    \dvq{q}_l = \vect{q}_c + \vect{T}_l \epsilon \\
    \dvq{\xi}_l = \dvqd{q}_l = \vectd{q}_c + \vectd{T}_l \epsilon,
\end{gather}
where $\vect{q}_c$ is the direction vector from the UAV to the load and $\vect{T}_l$ is the position of the load in the initial frame. 

\subsection{Dynamic Model with Slack Cable}
\label{subsec:slack}
When the cable is slack, the UAV and the sling load can be considered as two separate subsystems. 
The kinematics and dynamics of the UAV can refer to the rigid body model formulation in Sec.~\ref{subsec:dual dynamics}. 
For the sling load subsystem it is static on the ground or in free fall. 
The motion equations of the system are as follows
\begin{gather}
    \label{eq:wv slack}
    J_v\vectd{\omega}^b_v + \vect{\omega}^b_v \times (J_v\vect{\omega}^b_v) = \vect{\tau}_v \\
    \label{eq:av slack}
    m_v \vectdd{T}^b_v = \vect{F}_v \\
    \label{eq:al slack}     
    m_l \ddot{\vect{T}}_l = -m_l g\vect{e}_3 ,
\end{gather}
Combining the dual quaternion dynamic model presented in Sec.~\ref{subsec:dual dynamics},
the dynamic model in this case is
\begin{align}
\label{eq:slack}
    \dot{x} =& \left[ \begin{array}{c}
         \dqd{q}_v  \\
         \dqd{\xi}^b_v \\
         \dqd{q}_l  \\
         \dqd{\xi}^b_l
    \end{array}\right] = \left[ \begin{array}{c}
         \frac{1}{2} \dq{q}_v \otimes \dq{\xi}^b_v \\
         \dq{F}_v + \dq{u}_v  \\
         \vectd{T}_l\epsilon \\
         -g\vect{e}_3 \epsilon
    \end{array}\right] 
\end{align}
\subsection{Dynamic Model with Taut Cable}
\label{subsec:taut}
When the cable is taut, the conﬁguration of the system is deﬁned by the position and attitude of the load with respect to the inertial frame, and the UAV attitude. 
The position and velocity relationships between the UAV and the load are expressed as
\begin{align}
    \label{eq:Tv and Tl}
    \vect{T}_v = \vect{T}_l +  l\vect{q}_c \\
    \label{eq:vv and vl}
    \vectd{T}_v = \vectd{T}_l +  l\vectd{q}_c 
\end{align}
where $\vect{T}_v$ and $\vect{T}_l$ are the positions of the UAV and the load, and $l$ is the length of the cable between the UAV and the load.

The motion equations of the sling load are obtained from previous work~\cite{sreenath2013geometric,Pries2021}.
The motion equations of the cargo UAV system are
\begin{gather}
    \label{eq:wv_dot}
    J_v\vectd{\omega}^b_v + \vect{\omega}^b_v \times (J_v\vect{\omega}^b_v) = \vect{\tau}_u + \vect{\tau}_\text{ext} \\
    \label{eq:Tl_ddot}
    m \vectdd{T}_l + m_v l (\vectd{q}_{c}\cdot \vectd{q}_{c})\vect{q}_{c} = (\vect{q}_{c} \cdot \vect{F}_{uI})\vect{q}_{c} + \vect{F}_\text{ext}\\
    \label{eq:qc_ddot}
    m_v l \vectdd{q}_c + m_v l (\vectd{q}_{c}\cdot \vectd{q}_{c})\vect{q}_{c} = \vect{q}_{c} \times \vect{q}_{c} \times \vect{F}_{uI},
\end{gather}
with $m = m_v + m_l$ as the sum of UAV's mass $m_v$ and load's mass $m_l$. $\vect{F}_{uI} = \quat{q}_{v} \otimes \vect{F}_{th} \otimes \quat{q}_{v}^{*}$ is the expression of thrust vector $\vect{F}_{th}$ of the UAV  in the initial frame.
The full dynamical model has then obtained according to Sec.~\ref{subsec:dual dynamics}
\begin{align}
\label{eq:nonzero x}
    x = \left[ \begin{array}{c}
         \dvq{q}_l  \\
         \dvq{\xi}_l \\
         \dq{\xi}^b_v 
    \end{array} \right], \dot{x} = \left[ \begin{array}{c}
         \dvqd{q}_{l}  \\
         \dvqd{\xi}_{l} \\
         \dqd{\xi}^b_v 
    \end{array}\right] = \left[ \begin{array}{c}
         \vectd{q}_c + \vectd{T}_l \epsilon  \\
         \dq{\vect{F}}_l + \dq{\vect{u}}_l \\
         \dq{F}_v + \dq{u}_v 
    \end{array}\right]
\end{align}
with
\begin{align}
    \label{eq:uv}
    &\dq{u}_v = J_v^{-1}\vect{\tau}_u + [J_v^{-1}\vect{\tau}_u \times \vect{T}_{v}^b  + \vectdd{T}_{v}^b]\epsilon  \\ 
    \label{eq:Fv}
    &\dq{F}_v = a_v + [a_v \times \vect{T}_{v}^b + \vect{\omega}^b_v \times \vectd{T}_v^b]\epsilon   \\ 
    \label{eq:av}
    &a_v = J_v^{-1}(\vect{\tau}_\text{ext} - \vect{\omega}^b_v \times J_v \vect{\omega}^b_v )      \\
    \label{eq:ul}
    &\dq{\vect{u}}_l = \frac{ \vect{q}_c \times \vect{q}_c \times \vect{F}_{uI}}{m_v l} + \frac{(\vect{q}_c \cdot \vect{F}_{uI}) \vect{q}_c}{m}\epsilon  \\
    \label{eq:Fl}
    &\dq{\vect{F}}_l = -(\vectd{q}_c\cdot \vectd{q}_c)\vect{q}_c + [\frac{\vect{F}_\text{ext} - m_v l (\vectd{q}_c\cdot \vectd{q}_c)\vect{q}_c}{m}]\epsilon.
\end{align}

\section{Control Design for Cargo UAV System}
\label{sec:4 Control design}
\label{subsec:takeoff}
\begin{figure}[tp]
    \centering
    \includegraphics[width=7cm]{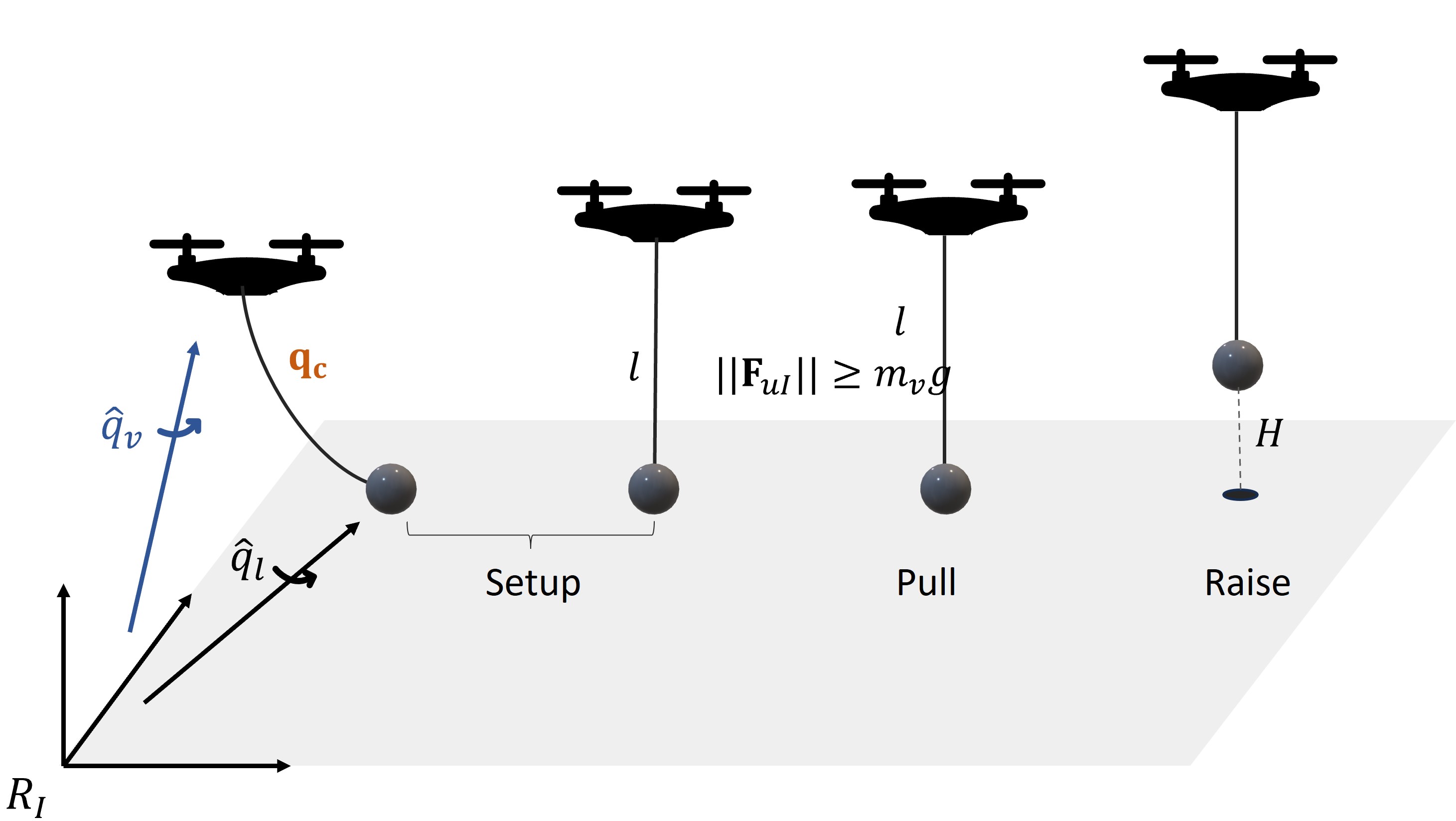}
    \caption{The lifting maneuver of the cargo UAV is separated into three modes: \textit{Setup},  \textit{Pull}, and  \textit{Raise}.}
    \label{fig:lift}
\end{figure}
\subsection{Lifting Modes and Desired Motion States}
\label{subsec:Lifting Modes}
The lifting maneuver is sketched in Fig.~\ref{fig:lift}, which is broken into three modes: \textit{Setup}, \textit{Pull}, and \textit{Raise}. In this way, the dynamics of the cargo UAV system in specific regimes during the maneuver can be characterized separately according to the state of the cable. During the \textit{Setup} and \textit{Pull} modes, the makes preparations to lift the load, and in the \textit{Raise} mode, the load is finally lifted off the ground.

\textbf{\textit{Setup}:}
\label{item:Setup}
As shown in Fig.~\ref{fig:lift}, the UAV has to fly to the point above the load from its initial position. During the process, the cable is slack, so the system can be considered as two separate subsystems. The dynamics of the UAV and the load are given in Sec.~\ref{subsec:slack}, while the load remains static.

The UAV is expected to reach the point right above the load with the distance between them as the cable length. Then, the desired configuration of the UAV is given by $q_v^{d.s}$ and $\vect{T}_{v}^{d.s}$ in Table~\ref{tab:params}.

The cargo UAV system turns to \textit{Pull} mode when the distance between the UAV and the load is equal to the cable length $l$, which means the cable is fully extended. It can be expressed as $l = ||\vect{T}_{v} - \vect{T}_{l}||$. To avoid swinging when lifting the load, we also consider that the UAV should be in a stable condition, which is measured by the errors $(q_{ve}, \dq{\xi}^b_{ve})$ converging to $(I, \dvq{0})$.    

\textbf{\textit{Pull}:}
\label{item:Pull}
Even though the cable is not slack anymore after \textit{Setup} mode, the force on the cable is initially not sufficient to lift the load. Therefore, the load is still static in dynamic model~\eqref{eq:nonzero x}. We assume the cargo UAV turns into \textit{Raise} mode until $||\vect{T}_{v} - \vect{T}_{l}|| = l$ and $m_vg  \leq||\vect{F}_{uI}||$, where $l$ is the cable length.

\textbf{\textit{Raise}:}
\label{item:Raise and Tracking}
The UAV is ready to raise the sling load when $l = ||\vect{T}_{v} - \vect{T}_{l}||$. During this mode, the load is not static anymore. The motion equations for the cargo UAV system are as well coupled as described in Sec.~\ref{subsec:taut}.
From pulling until the end of raising, the load is expected to reach a desired height, which is given by $H$ in Table~\ref{tab:params}. The corresponding desired states are given by $\vect{T}_l^{d.r}$ and $\vectd{T}_l^{d.r}$. 

\subsection{Control Law for Slack Cable}
\label{subsec:control zero}
When the cable is slack during \textit{Setup}, the load remains static on the ground. We just consider the control of the UAV. A control law coming from~\cite{arrizabalaga2023pose,wang2013unit} is adopted to cancel the non-linearities of the UAV subsystem $\dq{\xi}^b_v = \dq{F}_v + \dq{u}_v$, and then we get
\begin{equation}
\label{eq:control zero}
    \dq{u}_v = -2\dvq{k}_{pv} \ln \dq{q}_{ve} - \dvq{k}_{vv} \dq{\xi}^b_{ve} - \dq{F}_v + E_{\dq{q}_{ve}^{*}}\dq{\xi}_v^{d}, 
\end{equation}
where $\dvq{k}_{pv} = \vect{k}_{pvr} + \vect{k}_{pvd} \epsilon$ and $\dvq{k}_{vv} = \vect{k}_{vvr} + \vect{k}_{vvd} \epsilon$ are control gains. Both of them are dual vector quaternions.

Combining Eq.~\eqref{eq:slack} and Eq.~\eqref{eq:control zero}, we obtain the control commands for the UAV 
\begin{align}
    \label{eq:tau_u}
        \vect{\tau}_u &= \vect{\tau}_v - \vect{\tau}_\text{ext} =-J_v (\vect{k}_{pvr}\overline{\vect{\theta}}_{ve} + \vect{k}_{vvr}\vect{\omega}^b_{ve} + a_v)  \\
    \label{eq:F_u}
        \vect{F}_u &= \vect{F}_v - \vect{F}_\text{ext} =-m_v [\vect{k}_{pvd}\vect{T}^b_{ve} + \vect{k}_{vvd}(\vect{\omega}^b_{ve} \times \vect{T}^b_{ve} \\
        & + \vectd{T}^b_{ve}) + (a_v \times \vect{T}^b_{v} + \vect{\omega}^b_v \times \vectd{T}^b_v)] - m_v \overline{g}  \nonumber.
\end{align}
As the assumptions mentioned in Sec.~\ref{subsec:assumptions}, we have the external torque  $\vect{\tau}_\text{ext} = 0$, and the external force $\vect{F}_\text{ext}=m_v \overline{g}$ with $\overline{g} = \quat{q}_v^* \otimes g\vect{e}_3 \otimes \quat{q}_v$ as the representation of the gravity in the body frame.

\subsection{Control Law for Taut Cable}
\label{subsec:control nonzero}
We aim to track a desired trajectory with the sling load while keeping the desired yaw angle of the UAV. The position of the sling load is governed by the control of the UAV and the direction vector of the cable.
We define the desired configuration and twist of the sling load as $\dvq{q}_l^d = \vect{q}_c^d + \vect{T}_l^d \epsilon$ and $\dvq{\xi}_l^d = \vectd{q}_c^d + \vectd{T}_l^d \epsilon$, 
where $\vect{T}_l^d$ and $\vectd{T}_l^d$ are the desired position and velocity of the load. $\vect{q}_c^d$ is the desired load attitude. The desired yaw angle of the UAV is set as zero.
From the dual vector quaternion error dynamics of the sling load, we have the tracking errors as
\begin{gather}
    \label{eq:dual_q_le}
    \dvq{q}_{le} = \vect{q}_{ce} + \vect{T}_{le}\epsilon \\
    \label{eq:dual_xi_le}
    \dvq{\xi}_{le} = \vectd{q}_{ce} + \vectd{T}_{le}\epsilon.
\end{gather}
where $\vect{T}_{le}$ and $\vectd{T}_{le}$ are the load position and velocity errors. $\vect{q}_{ce}$ and $\vectd{q}_{ce}$ are the errors of load attitude, which are expressed as 
\begin{align}
    \label{eq:qce}
    \vect{q}_{ce} =& \vect{q}_c \times (\vect{q}_c \times \vect{q}_c^d) \\
    \label{eq:qce_dot}
    \vectd{q}_{ce} =& \vectd{q}_{c} - (\vect{q}_c^d \times \vectd{q}_c^d) \times \vect{q}_c.
\end{align}

From the dynamic model~\eqref{eq:nonzero x}, the control force $\vect{F}_{uI}$ has coupling terms between the UAV's attitude and the load's position and attitude. Meanwhile, we can see that the UAV's attitude is determined by its desired yaw angle and the load's position and attitude. Therefore, a controller for the load's attitude and position is designed as Eq.~\eqref{eq:load controller}, and a controller for the UAV's attitude is designed considering the dual quaternion error dynamics described in~\ref{subsec:error dynamics}
as Eq.~\eqref{eq:drone controller}.
\begin{align}
    \label{eq:load controller}
    \dvq{u}_l &= -(\dq{\vect{k}}_{pl} \dvq{q}_{le} + \dq{\vect{k}}_{vl} \dvq{\xi}_{le} + \dvq{F}_l) \\ 
    \label{eq:drone controller}
    \vect{\tau}_u &= -J_v (\dvq{k}_{pv}\overline{\vect{\theta}}_{ve} + \dvq{k}_{vv}\vect{\omega}_{ve} + a_v)
\end{align}
where $\dvq{k}_{pl} = \vect{k}_{plr} + \vect{k}_{pld} \epsilon$, $\dvq{k}_{vl} = \vect{k}_{vlr} + \vect{k}_{vld} \epsilon$, $\vect{k}_{pv}$, and $\vect{k}_{vv}$ are control gains.

The desired attitude of the sling load is determined by the tracking errors and the desired acceleration, which is calculated by 
\begin{align}
    \label{eq:q_cd}
     \vect{q}_c^d = \frac{-\vect{k}_{pld}\vect{T}_{le} - \vect{k}_{vld}\vectd{T}_{le} + \vectdd{T}_l^d + g\vect{e}_3}{||-\vect{k}_{pld}\vect{T}_{le} - \vect{k}_{vld}\vectd{T}_{le} + \vectdd{T}_l^d + g\vect{e}_3||} 
\end{align}

Combining Eqs.~\eqref{eq:nonzero x},~\eqref{eq:load controller}, and~\eqref{eq:q_cd} we obtain that the desired control force satisfies
\begin{gather}
    \label{eq:F_u dot}
        (\vect{q}_{c} \cdot \vect{F}_{uI})\vect{q}_{c} = -m(\vect{k}_{pld}\vect{T}_{le} + \vect{k}_{vld}\dot{\vect{T}}_{le} + \dvq{F}_l^d)  \\ 
    \label{eq:F_u cross}
        \vect{q}_{c} \times \vect{q}_{c} \times \vect{F}_{uI} =  -m_v l (\vect{k}_{plr}\vect{q}_{ce} + \vect{k}_{vlr}\vectd{q}_{ce}  + \dvq{F}_l^r),
\end{gather}
where $\dvq{F}_l^r$ and $\dvq{F}_l^d$ indicate the real part and the dual part of $\dvq{F}_l$ respectively. Then, we obtain the desired $\vect{F}_{uI}$
\begin{gather}
\label{eq:F_uI}
    \vect{F}_{uI}=  (\vect{q}_{c} \cdot \vect{F}_{uI})\vect{q}_{c} - \vect{q}_{c} \times \vect{q}_{c} \times \vect{F}_{uI} \\
    = -m(\vect{k}_{pld}\vect{T}_{le} + \vect{k}_{vld}\dot{\vect{T}}_{le} + \hat{F}_l) + m_v l (\vect{k}_{plr}\vect{q}_{ce} + \vect{k}_{vlr}\vectd{q}_{ce}). \nonumber
\end{gather}

It is important to note that the desired control force $\vect{F}_u$ of the UAV in body frame that is necessary to move the load to the desired reference is given in Eq.~\eqref{eq:Tl_ddot} is referenced to the initial frame. $\vect{F}_u$ is obtained by quaternion product
\begin{equation}
\label{eq:F_u}
    \vect{F}_u = \quat{q}_{v}^{*} \otimes \vect{F}_{uI} \otimes \quat{q}_{v}.
\end{equation}

Considering that the quadrotor is an under-actuated system, the actual force exerted by the motors is given by
\begin{equation}
    \vect{F}_{th} = [
         0~
         0~
         ||\vect{F}_u|| 
         ]^T = [
         0 ~
         0 ~
         1
         ]^T ||\vect{F}_u|| = \vect{b} ||\vect{F}_u||
\end{equation}
For the attitude of the UAV, a quaternion $\quat{q}_t$ exists that rotates the UAV such that the thrust force vector $\vect{F}_{th}$ coincides with the desired control force $\vect{F}_{uI}$ in the inertial frame, and it is defined as
\begin{equation}
    \quat{q}_t = \frac{(\vect{b} \cdot \vect{F}_{uI} + ||\vect{F}_{uI}||) + \vect{b} \times \vect{F}_{uc}}{||(\vect{b} \cdot \vect{F}_{uI} + ||\vect{F}_{uI}||) + \vect{b} \times \vect{F}_{uc}||}
\end{equation}
where $\vect{F}_{uc}$ is the current thrust force vector with respect to the inertial frame.
Then if the desired rotation around the $z$ axis is defined with a quaternion $\quat{q}_{zd}$, then $\quat{q}_t$ is completed using a quaternion product~\cite{abaunza2017dual}
\begin{equation}
\label{eq:qvd}
    \quat{q}_v^d = \quat{q}_{zd} \otimes \quat{q}_t
\end{equation}

Till now, $\quat{q}_v^d$ is used as the complete attitude reference to feedback the rotational control input from Eq.~\eqref{eq:drone controller} and $\vect{F}_{th}$ dictates the thrust driven by the rotors.

\begin{table}[!htbp]
    \centering
    \caption{Simulation Parameters of the cargo UAV system\tablefootnote{{Parameters in this paper, such as $J$, $m$, $\vect{T}$ ($l$), and $\vect{\omega}$, are all using International System of Units (\si{kg*m^2}, \si{kg}, \si{m}, and \si{rad/s})}. The units are omitted therein.} }
    \begin{tabular}{c||p{6cm}}
     \toprule[1pt]
    UAV     & $m_v = 0.7$ , $J_v = \textit{diag}[0.005, 0.007, 0.006]$, $\vect{d} =[0,l/2, 0, 0]$ $\vect{T}_{v0} = \vect{T}_{l0} + \vect{d}$, $\vectd{T}_{v0} = \vect{0}$, $q_{v0} = I$\\
    \midrule
    Load     & $m_l = 0.05$, $l=0.3$, $\vect{T}_{l0} = \vect{0}$, $\vectd{T}_{l0} = \vect{0}$\\
    \midrule
    Controller & $\dvq{k}_{pv} =[10, 10, 10] + [1, 1, 4]\epsilon$,$ \dvq{k}_{vv} = [1, 1, 1] + [1, 1, 4]\epsilon$, $\dvq{k}_{pl} = [2, 2, 2] + [1, 1, 4]\epsilon$, $\dvq{k}_{vl} = [0.5, 0.5, 0.5] + [1, 1, 4]\epsilon$ \\
    \midrule
    Trajectory & $q_v^{d.s} = I$, $\vect{T}_{v}^{d.s} = [0,0,0,l]$, $H = 3l$, $\vect{T}_{l}^{d.r} = H$, $\vectd{T}_{l}^{d.r} = [0,0,0.5\sin(\frac{\pi}{3}))]$, $t \in [0,6]$, $\vect{T}_{l}^{d.t} = [\sin(\frac{\pi}{6}), 0.5\sin(\frac{\pi}{6}), \max(0, 0.5\sin(\frac{\pi}{3}))]$, $t \in [0,6]$\\
    \bottomrule
    \end{tabular}
    \label{tab:params}
\end{table}

\section{Simulation}
\label{sec:5 simulation}
\subsection{Simulation Settings}
\label{subsec:settings}
The performance of the proposed control strategy is tested in a set of numerical simulations with a time step of \SI{0.01}{s}.  
The parameters of the cargo UAV system, the control gains, initial values, and parameters for the desired trajectories are depicted in Table~\ref{tab:params}.
\begin{figure}
    \centering
    \includegraphics[width=0.52\textwidth]{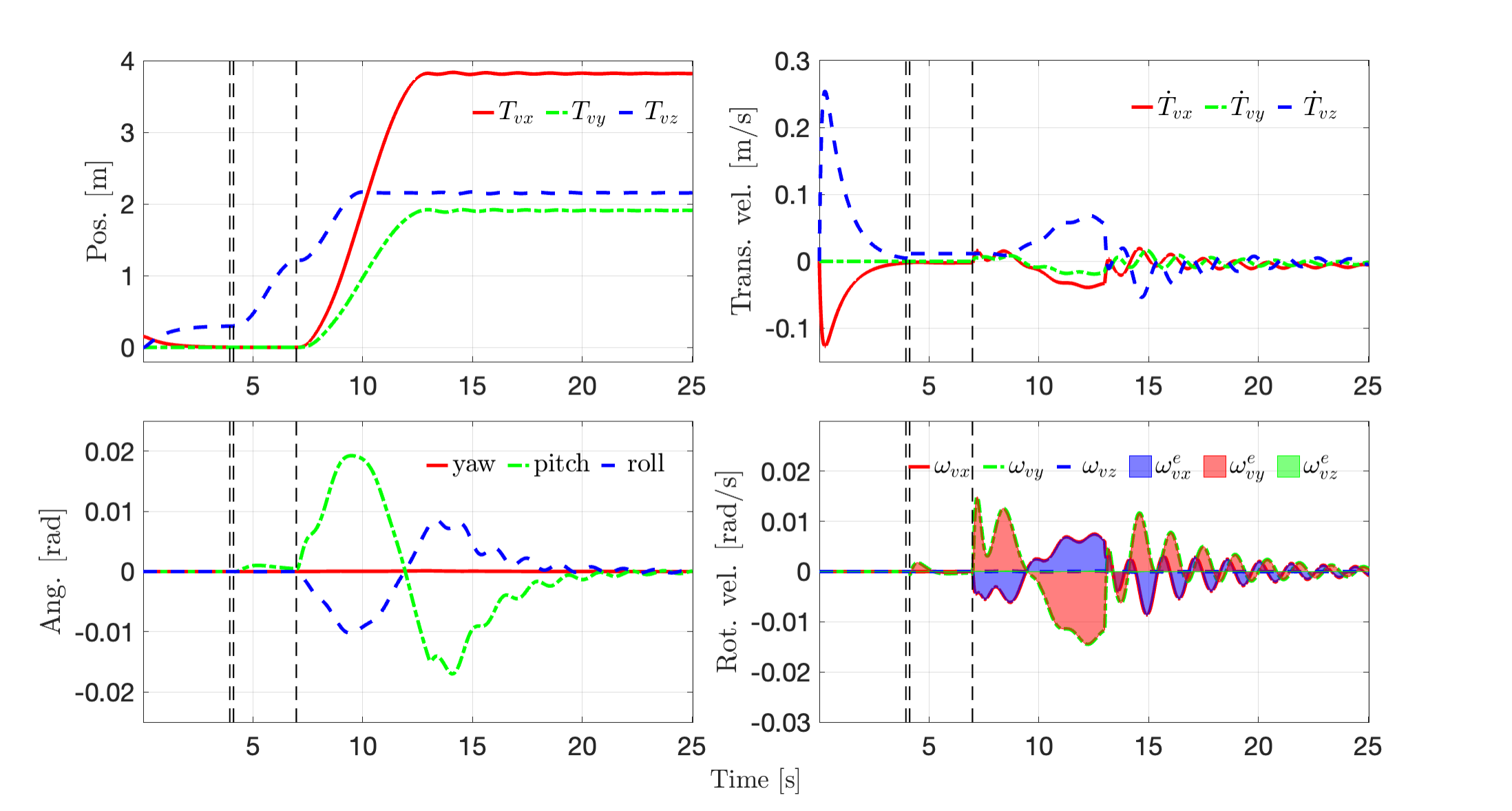}
    \caption{States of the UAV during the lifting and trajectory tracking process. In these figures, the vertical black dashed lines indicate the time at which the UAV switches the mode to the next one, colored lines show the real values and shade areas are the errors with the desired values. The first figure depicts the position of the UAV and the second one presents the translational velocity. The two figures in the bottom row show the attitude and angular velocity of the UAV. } 
    \label{fig:drone}
\end{figure}
\subsection{Simulation 1: Lifting and Trajectory Tracking}
\label{subsec:sim 1}
In this simulation, we conduct the tracking of a trajectory including the take-off maneuver and the tracking of a trajectory with the load. The sling load lifting maneuver is simulated following the three modes~\cite{cruz2017geometric} illustrated in Fig.~\ref{fig:lift} and explained in Sec.~\ref{subsec:Lifting Modes}. After the sling load reaches a desired height, the reference trajectory is generated and the sling load tracks the trajectory. The desired trajectory is determined by $\vectd{T}_{l}^{d.t}$ in Table~\ref{tab:params}.

\begin{figure}
    \centering
\includegraphics[width=0.52\textwidth]{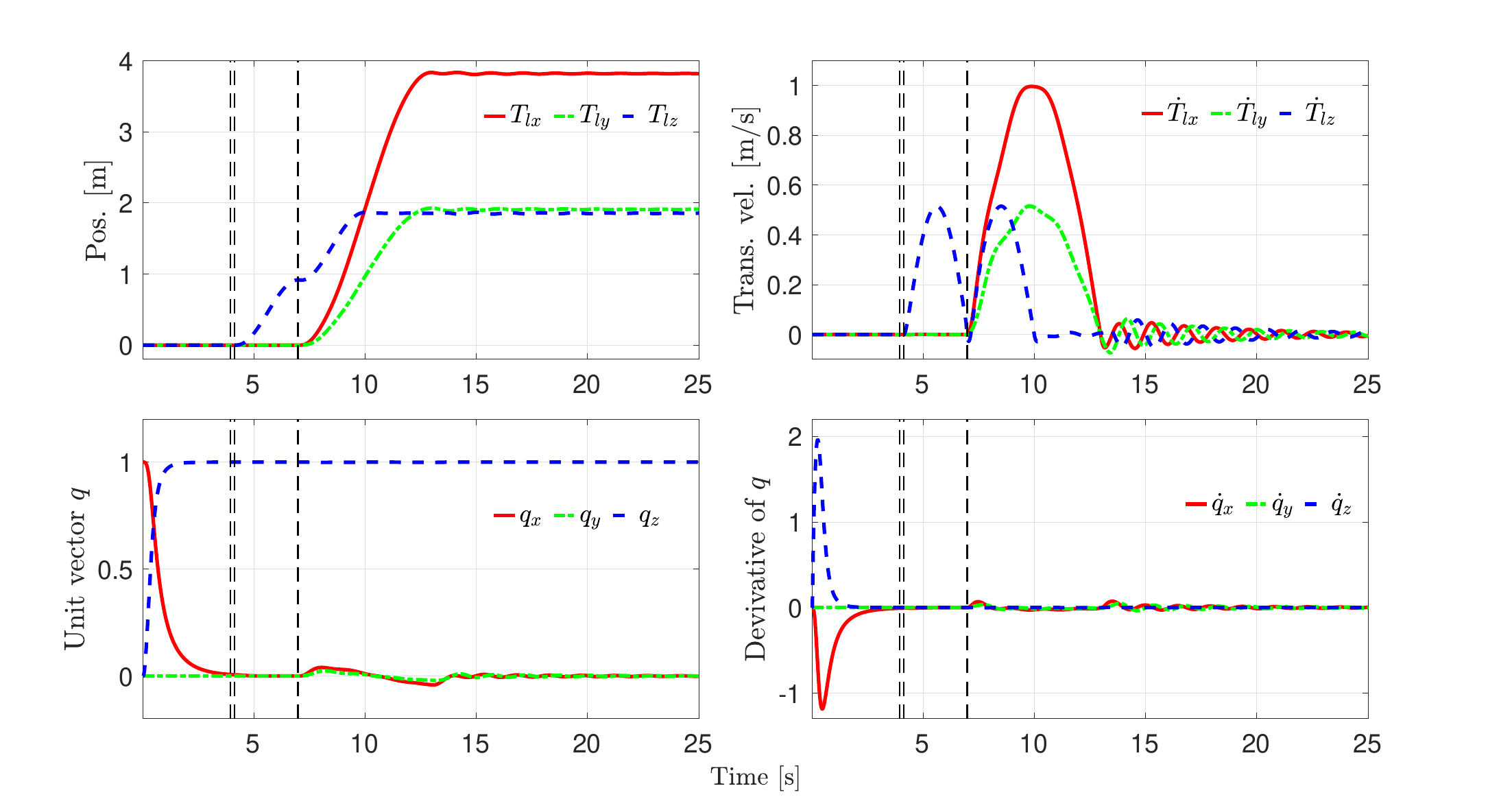}
    \caption{States of the load during the lifting and tracking process. The vertical black dashed lines indicate the time at which the UAV switches the mode to the next one. The two figures in the first row show the position and translational velocity of the load. The figures in the bottom row depict the unit direction vector $\vect{q}_c$ and its derivative.} 
    \label{fig:load}
\end{figure}

\begin{figure}
    \centering
    \includegraphics[width=0.52\textwidth]{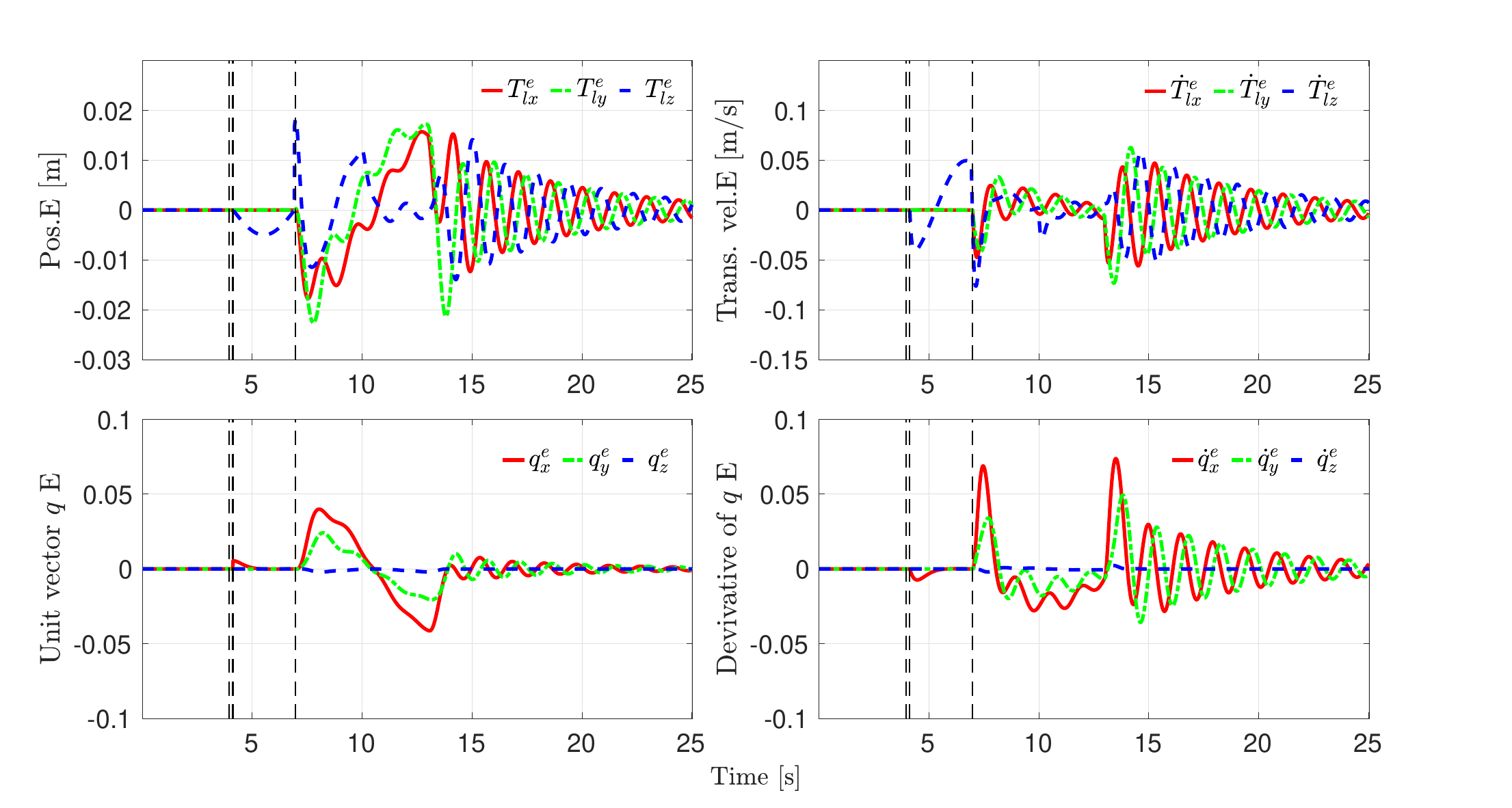}
    \caption{State errors of the load during the lifting and tracking process. The vertical black dashed lines indicate the time at which the mode jumps from one to the next. The two figures in the first row show the position and translational velocity errors of the load. The figures in the bottom row depict the errors for the unit direction vector $q_c$ and its derivative.} 
    \label{fig:error}
\end{figure}

The simulation results are depicted in Fig.~\ref{fig:drone}, Fig.~\ref{fig:load}, and Fig.~\ref{fig:error}.
The UAV takes about \SI{3.9}{s} to finish the setup process.
Then, in a very short time, the UAV completes the \textit{Pull} mode (about \SI{0.2}{s}) being in a stable state avoiding cable oscillation~\cite{bisgaard2009slung}.
The total time of accomplishing the load lifting maneuver takes $\approx$\SI{6.9}{s}. 
The results show that the UAV maintains stable status above the load throughout the \textit{Setup} and \textit{Pull} phases. During these phases, the unit direction vector $\vect{q}_c$ adapts to the UAV's position, as depicted in Fig.~\ref{fig:load}. Upon entering the \textit{Raise} mode, the load ascends from the ground and approaches the desired height. Notably, the errors $\vect{q}_{ce}$ and $\dot{\vect{q}}_{ce}$ in this mode remain close to zero, indicating minimal swing of the sling load, as shown in Fig.~\ref{fig:error}. 

During the tracking maneuver, the position and translational velocity of the UAV are determined by the position of the sling load and the direction vector $\vect{q}_c$, as shown in Fig.~\ref{fig:drone}. The attitude of the UAV is obtained from Eq.~\eqref{eq:qvd}. Combining the unit vector $\vect{q}_c$ and $\vect{q}_c^d$ shown in Fig.~\ref{fig:load}, we can see that the attitude of the UAV changes when the sling load tracks the reference trajectory. After the tracking task is completed, the attitude of the UAV stabilizes horizontally with minimal tracking errors. 

The state errors of the sling load are depicted in Fig.~\ref{fig:error}. The position errors in the first figure indicate the sling load is able to track the desired trajectory with a maximum error around \SI{-0.02}{m}. Combining the translational velocity in the second figure, we can see the developed dual quaternion controller performs well for the sling load lifting and tracking tasks. The errors of $\vect{q}_c$ and $\dot{\vect{q}}_c$, shown in the second row, indicate that the proposed control strategy is able to minimize the swing of the load during the maneuver sufficiently. 

\subsection{Simulation 2: System Evaluation}
\label{subsec:sim 2}
\begin{figure}\includegraphics[width=.5\textwidth]{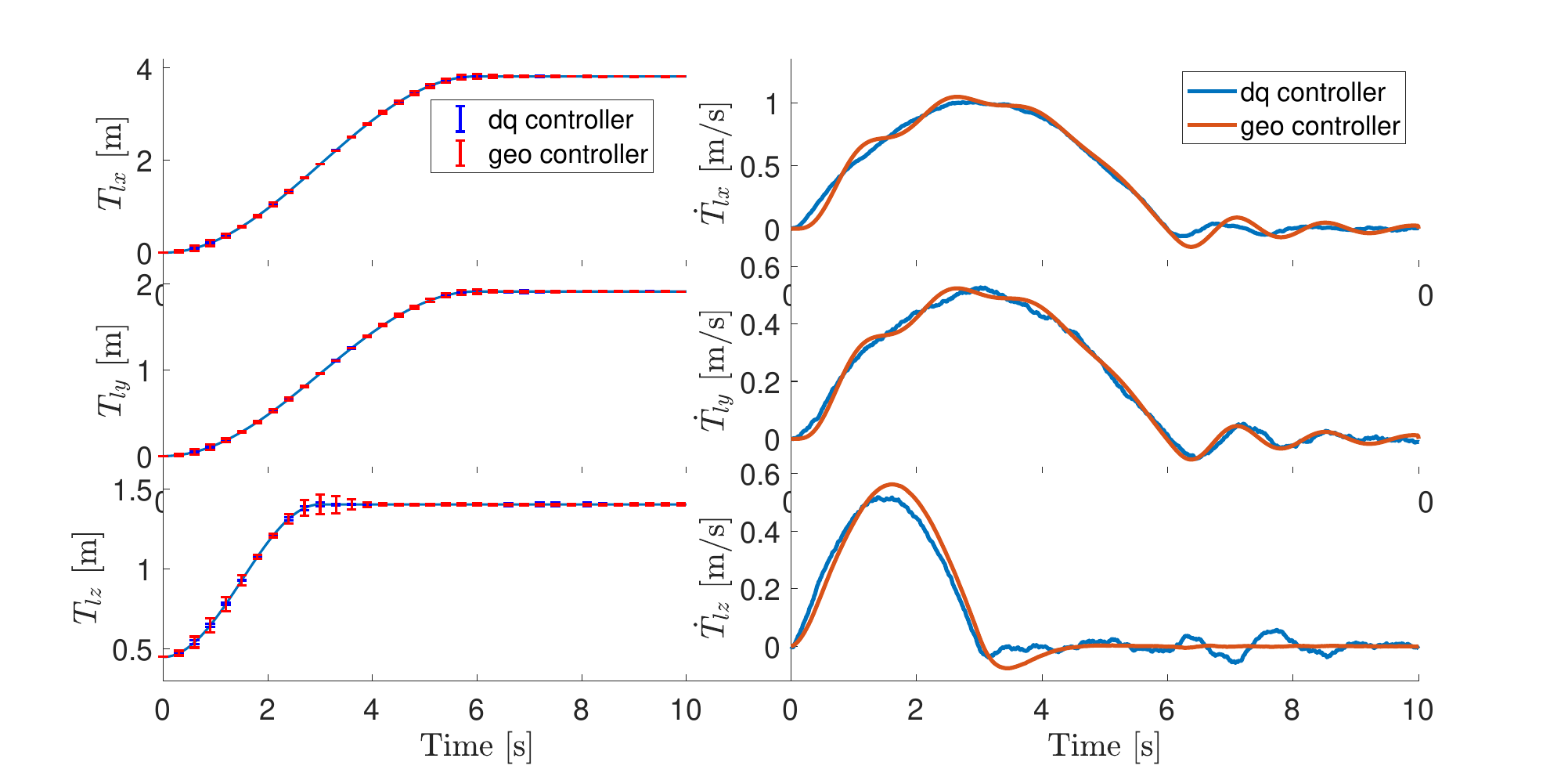}
    \caption{The comparison results of our dq (dual quaternion) controller and the geo (geometric) controller. The left figure shows the distribution of tracking errors, and the right one depicts the velocity. } 
    \label{fig:comparison}
\end{figure}
\begin{figure}\includegraphics[width=.5\textwidth]{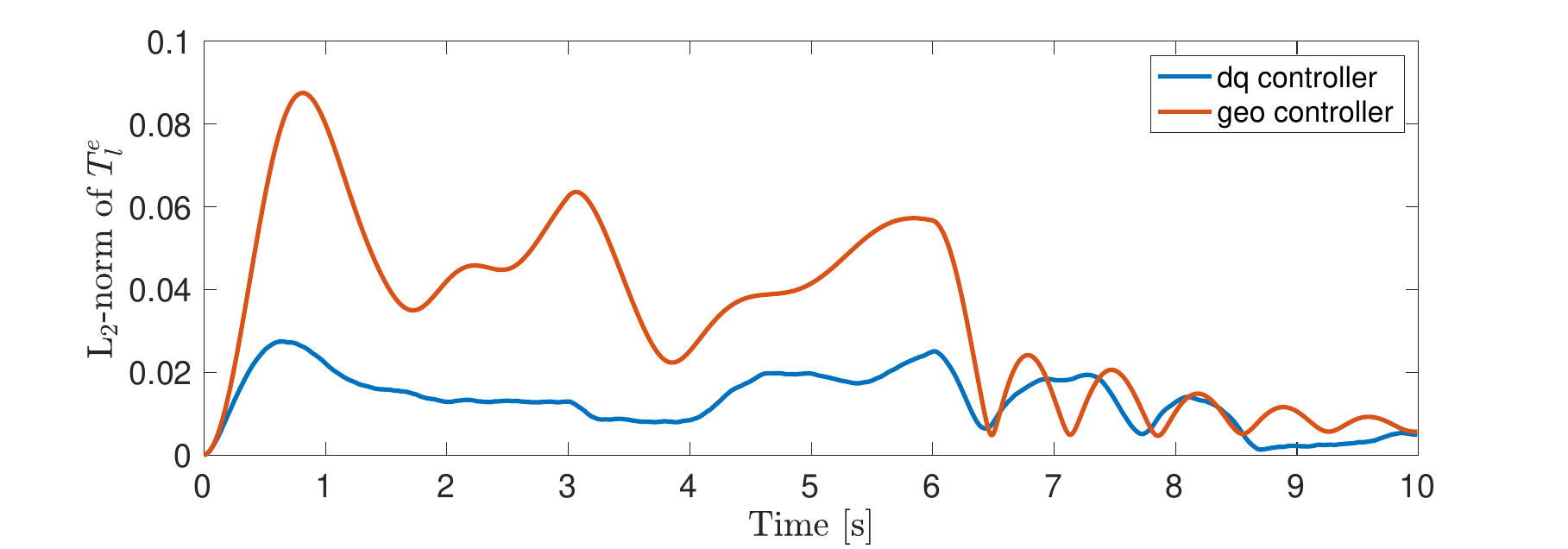}
    \caption{The L$_2$-norm of tracking errors for the proposed dq (dual quaternion) controller and the geo (geometric) controller. } 
    \label{fig:norm_e}
\end{figure}
The purpose of this set of simulations is to test the robustness of the proposed controller. To do that we considered both, external disturbances for control inputs, forces $\vect{F}$ and torques $\vect{\tau}$, and system uncertainties. For system uncertainties, we focused on system parameters that can not be measured precisely including $m_v$, $m_l$, $l$, and $J_v$.
In our simulation, we added white Gaussian noise to these variables with the signal-to-noise ratio of $35$. 
For comparison of the performance of our proposed model, we implemented the geometric controller presented in~\cite{sreenath2013geometric} with the same settings as the baseline. 
The simulation was conducted $100$ times for the two controllers and each time these parameters were initialized randomly by adding noise to the ideal values.
The comparison results are shown in Fig.~\ref{fig:comparison} and Fig.~\ref{fig:norm_e}.

The left figure in Fig.~\ref{fig:comparison} presents the average tracking errors for $100$ times simulations. The right one depicts the corresponding velocity.
From the left figure, we can see the tracking errors of both our controller and the geometric controller are very small. However, the norm of tracking errors shown in Fig.~\ref{fig:norm_e} indicates that our controller performs better than the geometric control. This conclusion is also supported by the velocity plots Fig.~\ref{fig:comparison}. The dual quaternion control yields smooth velocity control. The reason for the fluctuation happening around \SI{7}{s} might be the sudden changes of velocities on the x-axis and y-axis, and the effects of control gains.  
These results demonstrate that the sling is able to track a desired trajectory with small tracking errors and minimum swing using our developed controller.

\section{Conclusion}
\label{sec:6 Conclusion}
By utilizing the mathematical efficiency and compactness of dual quaternions, this study introduces a novel control strategy, where the kinematics and dynamics of the UAV and its sling load are represented by dual quaternions. This representation significantly improves the control performance of the cargo UAV system. 
Two simulations were designed to assess the proposed control strategy. The lifting and tracking maneuver simulation results show that the sling load is able to track a desired trajectory with small tracking errors and swings. The second simulation indicates that the proposed control strategy is robust to system uncertainties and disturbances. The comparison results with the geometric controller show that our controller is more stable and performs better for tracking tasks. 
In future work, we will consider a potential offset between the attachment point of the load and the CoM of the UAV. Then, we will test the proposed controller on real UAVs.

\bibliographystyle{IEEEtran}
\bibliography{ref}

\end{document}